\documentclass[sigconf, review=false,screen=true, nonacm =False, standardpagestyle=True]{acmart}
\usepackage{graphicx}
\usepackage{subcaption}
\usepackage{comment}
\usepackage{multirow}
\usepackage{multicol}
\usepackage{xcolor}

\AtBeginDocument{%
  \providecommand\BibTeX{{%
    \normalfont B\kern-0.5em{\scshape i\kern-0.25em b}\kern-0.8em\TeX}}}

\setcopyright{rightsretained}
\copyrightyear{2021}
\acmYear{2021}
\acmDOI{}
\acmISBN{} 

\acmConference[KDD Workshop on Data-driven Humanitarian Mapping, 27th ACM SIGKDD Conference]{ August 15}{2021}{Virtual Conference}

\begin{document}

\title[What a million Indian farmers say?]{What a million Indian farmers say?: A crowdsourcing-based method for pest surveillance }

\author{Poonam Adhikari}
\email{poonam.19csz0007@iitrpr.ac.in}
\affiliation{%
  \institution{Indian Institute of Technology Ropar}
  \city{Ropar}
  \country{India}
  }

\author{Ritesh Kumar}
\affiliation{%
  \institution{CSIR-Central Scientific Instruments Organisation}
  \streetaddress{Sector 30 C}
  \city{Chandigarh}
  \country{India}}
\email{riteshkr@csio.res.in }

\author{S.R.S Iyengar}
\affiliation{%
  \institution{Indian Institute of Technology Ropar}
  \city{Ropar}
  \country{India}
}
\author{Rishemjit Kaur}
\affiliation{%
  \institution{CSIR-Central Scientific Instruments Organisation}
  \streetaddress{Sector 30 C}
  \city{Chandigarh}
  \country{India}}
\email{rishemjit.kaur@csio.res.in}
\renewcommand{\shortauthors}{Adhikari and Kaur, et al.}

\begin{abstract}
Many different technologies are used to detect pests in the crops, such as manual sampling, sensors, and radar. However, these methods have scalability issues as they fail to cover large areas, are uneconomical and complex. This paper proposes a crowdsourced based method utilising the real-time farmer queries gathered over telephones for pest surveillance. We developed data-driven strategies by aggregating and analyzing historical data to find patterns and get future insights into pest occurrence. We showed that it can be an accurate and economical method for pest surveillance capable of enveloping a large area with high spatio-temporal granularity. Forecasting the pest population will help farmers in making informed decisions at the right time. This will also help the government and policymakers to make the necessary preparations as and when required and may also ensure food security. 
\end{abstract}




\maketitle

\section{Introduction} 
Agriculture accounts for 20.19\% of India's GDP \cite{IndiaGDP22:online}, and on average, it employs over 50\% of the population as per Indian economic survey \cite{Indiaeco86:online}. Over the past three decades, India achieved remarkable growth in agriculture production by expanding agricultural land, usage of high-yielding varieties of crops, fertilizers, and pesticides. However, increasing production through the expansion of farming areas is limited. Besides, production is restrained by several other factors such as limited availability of irrigation water, erratic weather patterns, lowering of groundwater levels, decreasing fertile land (availability per capita), and potential damage to crops due to pests.

Pests have been responsible for extensive crop damage in countless ways affecting overall production. Irrespective of the crop, pest attacks have been found in roots, stalks, bark, stems, leaves, buds, flowers, and fruits of plants \cite{atwal2015agricultural}. According to the global burden \cite{finegold2019global} of crop loss, pests and diseases cause about 20-40\% loss in agriculture. In India, pests cause an annual loss of about US \$42.66 million in agriculture \cite{sushil2016emerging}, implying pest infestation is a serious problem.  Although this problem persists for a long time, our efforts to control and manage the damage have not been entirely resolved. Understanding pest population dynamics is an effective step towards pest management that can reduce harmful pesticides. However, monitoring pests population is complex as their population fluctuates with time, crop, location, and season. Therefore, a precise understanding of population dynamics is a challenging task. 
\begin{table*}[!ht]
\centering 
\begin{tabular}{|p{10mm}|p{20mm}|p{18mm} |p{30mm}|p{25mm}|p{25mm}|} 
\hline
\textbf{Season} & \textbf{Sector} & \textbf{Category} & \textbf{Crop} & \textbf{Query Type}  &\textbf{ Query Text} \\ \hline
RABI & AGRICULTURE & Pulses & Black Gram (urd bean) & Plant Protection & pod borer in black gram \\ \hline
\multicolumn{2}{|p{20mm}|}{\textbf{KCC Ans}} & \textbf{State Name} & \textbf{District Name} &\textbf{Block Name} &  \textbf{Created On}   \\ \hline
\multicolumn{2}{|p{30mm}|}{recommended for spray quinalphos 2ml /liter}  & TAMILNADU & TIRUCHIRAPPALLI & PULLAMBADI & 2015-03-14 15:35:05.087 \\ \hline
\end{tabular}
\caption{An example of Kisan Call Center (KCC) data. It consists of 11 different fields.} 
\label{table:Kcc_data_format}
\vspace{5mm}%
\end{table*}

\begin{figure*}[!ht]
  \includegraphics[width=0.9\linewidth]{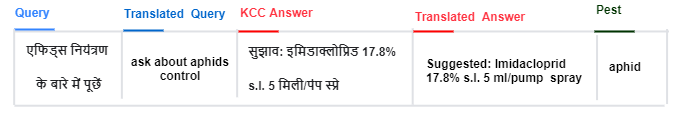}
  \caption{Example of KCC query and answer provided by KCC executive}
  \label{query_eg}
\end{figure*}

Traditionally, pest management practises include usage of pesticides such as fungicides, herbicides, and insecticides. However, these are toxic to human health and are responsible for contaminating the agriculture ecosystem \cite{BILAL2019133896}. Over the years, researchers have employed different sensors and IOT based methods for pest detection and monitoring. Some of these methods include usage of acoustic sensors, radars, LiDAR \cite{dwivedi2020insect}, cameras\cite{turkouglu2019plant}, infrared sensors, etc  \cite{saranya2019iot}. However, these methods lack in scalability due to low area coverage by sensors and are also expensive to deploy. Satellite imagery couple with machine learning techniques has also been used for pest detection \cite{yuan2014damage}. Another approach includes pest forecasting based on the weather and topological information. Wu et al. \cite{wu2008application} used multifactor spatial interpolation model which uses weather information as one of the data source and used geography information system (GIS) for forecasting. Hooker et al. \cite{hooker2002using} used multiple linear regression on rainfall, humidity and temperature data to predict pest in wheat plants. However, these studies focus on single plant or pest. There is a need for economical and scalable method for collecting high frequency and high resolution data.


In this work, we have proposed a crowdsourced data collection and pest attack prediction methodology. 
Crowdsourced data has been shown to be powerful tool in predicting infectious disease outbreak such as influenza and middle east respiratory syndrome \cite{seo2017methods} and dengue fever \cite{husnayain2019correlation}. It has the capability of capturing high frequency information keeping the cost low and provides wide spatial extent and high resolution. We have examined crowdsourced data recorded from farmer call centers, popularly know as `Kisan call centers' (KCC) \cite{KisanCal14:online} managed by Ministry of Agriculture, Government of India, which is available online at \cite{OpenGove38:online}. KCCs have been established throughout India to inform and advise farmers on the queries posed by them and proved to be efficient as 61.5\% of farmers depend on KCC for information support on various agricultural issues \cite{parihar2010sustainable}. They have been a successful model to establish direct communication between experts/semi-experts and the farmers. This big data consists of agricultural queries and can be used for investigating various problems faced by farmers \cite{nguyenbig}. Viswanath et al. \cite{viswanath2018hadoop} analyzed the KCC datasets to find the peak query hours. Mohapatra et al. \cite{mohapatra2018using} studied KCC data to answer most common queries by generating FAQ, Jain et al. \cite{jain2019agribot} have used KCC data to develop a chat bot for farmers. However, to the best of our knowledge no such work has been done in the field of pest prediction using KCC data. Our method is a data-driven approach to gather pest population information, which is accurate, economical, and is capable of covering large area with block-level granularity.

We used time-varying agricultural queries from different districts of India to study pest population dynamics. Interestingly, the spatio-temporal mapping of the normalized query features mimicked the events panning out in the real world accurately as examined by contemporary news articles and reports. Further, we developed machine learning models to forecast the pest population across India. This forecasting model will not only help in knowing pest arrival time but also identify the period of occurrence. This could help to track and predict pest severity and provide novel signals to the government or other agencies engaged in this area to preempt any outbreak.

\section{Methodology}

\subsection{Data}

Our dataset consists of KCC queries for 2015–2020 including 10,981,793 queries from 31 states and 553 districts collected from the Open Government Data (OGD) website \cite{OpenGove38:online}. It consists of season, crop category, query type, crop, the question asked by the farmer, and answer provided by the KCC executive, location information, and time of query raised. The query season reflects the harvest of the crop, Rabi, Kharif, or Zaid crops. Category field contains information about the query category, such as pulses, vegetables, etc.  The location is divided into three fields: state name, district name, and block name. An example of the data is shown in the Table \ref{table:Kcc_data_format}.

\subsection{Labelling of pest queries}
KCC data is not systematically organized and maintained leading to many challenges, such as missing information, spelling mistakes, and usage of code-mixed multiple languages. We performed data cleaning by discarding queries that have incomplete details, such as missing state names, district names, and creation dates. Further, we translated the queries to English using Google Translate API and spell corrected using TextBlob \cite{TextBlob74:online} python library for both question and KCC answer. An example has been shown in Figure \ref{query_eg}. 

\begin{figure*}[!ht]
  \includegraphics[width=\linewidth]{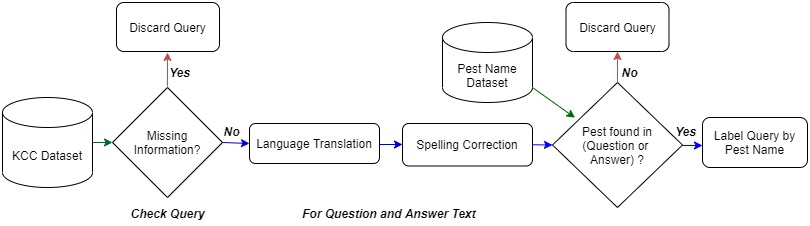}
  \caption{ Flowchart for labelling pest queries.}
  \label{steps}
  \vspace{-3mm}%
\end{figure*}

\subsection{Pest Seasonality and Forecasting}
In order to extract pest related queries, we generated a list of possible names for each pest by considering one character error and kept queries for which pest name is found either in question raised by a farmer or the answer given by the KCC executive. The flow chart labelling the pest queries is shown in Figure \ref{steps}. Following this procedure, we are left with a dataset of size 867,337 for pest related queries which is 7.89\% of total queries.
Also, we aggregated our data by date, district, and pest to calculate the frequency of daily pest queries from a specific district. As the pest frequency depends on the agricultural land, the pest frequency from a district was normalized by gross cultivable area \cite{ICRISATD46:online} in the given district.  


\begin{table}
\centering
\begin{subtable}[t]{0.48\textwidth}
\centering
\begin{tabular}{l|l|l|l|l} 
\hline 
\textbf{Result}&\textbf{Aphid}  & \textbf{Insect}& \textbf{Bug}& \textbf{Stemborer}\\ \hline
Test Statistic & -5.0867 & -4.2921 & -23.014&-3.217\\
p-value & 0.000015&0.00045&0.000&0.019\\
Log transformation & True & True & True& True \\
\#Lags Used & 10  & 13& 11&13\\
Critical Value (1\%)& -3.496& -3.496 &-3.501&-3.496\\
Critical Value (5\%)&-2.890 & -2.890 & -2.501&-2896\\
Critical Value (10\%)& -2.582&-2.582 & -2.583&-2.582\\
\hline
\end{tabular}
\caption{Results of Dickey-Fuller Test:  }
\label{table:dickey}
\end{subtable}
\hspace{\fill}
\begin{subtable}[t]{0.48\textwidth}
\centering

\begin{tabular}{l|l|l|l|l|l} 
\hline 

\multicolumn{2}{l|}{\textbf{Constant}}  &\textbf{Aphid} &  \textbf{Insect} & \textbf{Bug} & \textbf{Stemborer}\\ \hline
\multirow{3}{*}{AR} &p&1&1&2&1\\ \cline{2-6}
                    &d&0 &0&1&0\\ \cline{2-6}
                    &q&1&2&0&1\\  \hline

\multirow{3}{*}{SAR} &P&1&1&2&1\\ \cline{2-6}
                    &D&0& 0&1&1\\ \cline{2-6}
                    &Q&1&1&0&2\\ \cline{2-6}
                    &S&7& 12&11&8\\ \hline
\multicolumn{2}{l|}{AIC} &995.251&  942.906 &972.638 & 966.049 \\ \hline
\end{tabular}
\caption{SARIMA model parameters:}
\label{table:model_parameter}
\end{subtable}

\caption{ (a) The results of Dickey-Fuller test. Here, the parameters log transformation indicate whether log transformation is performed and \#lags used indicate the difference taken to make our data stationary. (b) The best fitted model parameters. Here, AR and SAR are autoregressive and seasonal autoregressive parameters, d and D are the trend difference and seasonal trend difference.}
\label{tab:sarima}
\vspace{-3mm}%
\end{table}
\begin{figure*}[!ht]
    \includegraphics[width=0.95\textwidth]{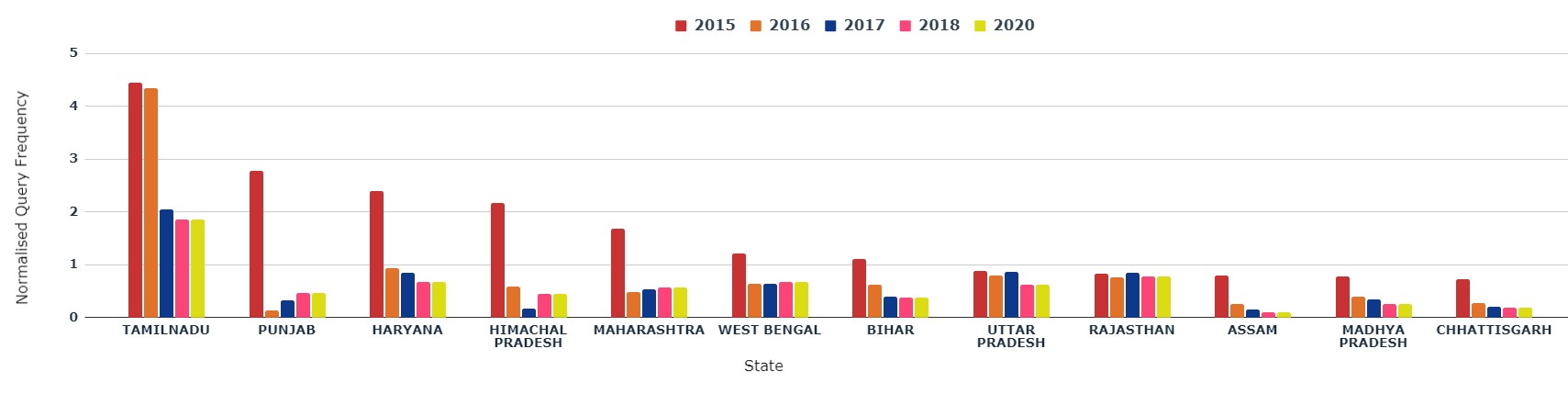}
    \caption{Figure showing normalized pest queries received from different states of India, for different years. Normalization is performed by dividing the frequency of queries from a given district by the gross cropped area in a given year.}
  \label{state_year}
  \end{figure*}

In order to quantify seasonality of pest population, we computed temporal auto-correlation to a lagged version of query frequency using \cite{bradley1968distribution}.
The auto-correlation vs. lag plot can be used to find seasonality in the pest attack. The periodicity of the peak shows the pattern after a period equal to the lag. 

Seasonal Auto-regressive Integrated Moving Average (SARIMA) is a forecasting method for continuous-time series data with seasonal variation. We used the SARIMA model for forecasting pest occurrences. So, as a first step we checked for data stationary property using Dickey-Fuller (DF) test \cite{mushtaq2011augmented}  as given by  \(y_i = \phi y_{i-1}+\epsilon_{i} \). Here, y$_i$ is the variable of interest, i is the time index, and $\epsilon$ is the error term.

For pest frequency time series where DF tests failed (i.e., non-stationary), we further take its log transformation followed by an iterative n-order differencing as shown in equation \ref{f_difference} till the stationarity has been achieved. To confirm the stationarity, we also checked for the time dependence of the mean and standard deviation. (i.e., rolling mean and rolling standard deviation are constant). Test statistics of the DF test is shown in Table \ref{table:dickey}
\begin{equation}
y_t = y_{t}-y_{t-n} 
\label{f_difference}
\end{equation} 

y$_t$ denotes the pest frequency at time $t$, and $n$ denotes seasonal difference. In order to determine the optimal values of SARIMA model parameters, i.e., trend autoregression order (p), the trend moving average order (q), seasonal autoregressive order (P), and seasonal moving average order (Q), we performed a grid search. Table \ref{table:model_parameter} shows the parameter combinations. Akaike information criterion (AIC) was used as a performance metric.

\section{Results}
\subsection{Pest frequency analysis}
We analyzed the five-year queries and found that the most significant number came from the agricultural sector at 79.42\%, followed by horticulture at 18.81\% and livestock, weather, and fishing below 1\%.  Among the different crops, cereals received the most queries with 20.9\% followed by pulses: 5.53\%, oil seeds: 4.82\%, fiber crops: 3.61\%, millets: 2.95\% and rest with less than 2\%.

We further analyzed only pest-related queries for 264 varieties of crops and observed that the top queries were related to paddy (dhan) with 16.2\% followed by cotton (kapas) 8.66\%, brinjal 7.62\%, sugarcane (noble cane) 4.45\%, wheat 4.29\% and rest below 3\%. We observe that Tamil Nadu receives the highest number of normalized pest queries as shown in  Figure \ref{state_year}.

\begin{figure}[!ht]
  \begin{subfigure}[b]{0.22\textwidth}
     \includegraphics[width=\linewidth]{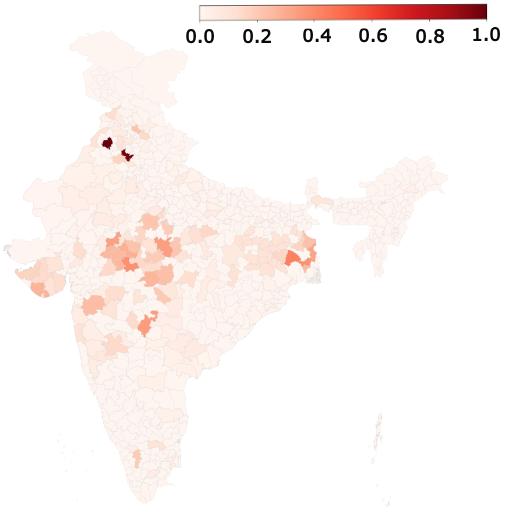}
  \caption{Whitefly queries during 2015.}
  \label{whitefly_news}
  \end{subfigure}
  \hfill
  \begin{subfigure}[b]{0.22\textwidth}
    \includegraphics[width=\textwidth]{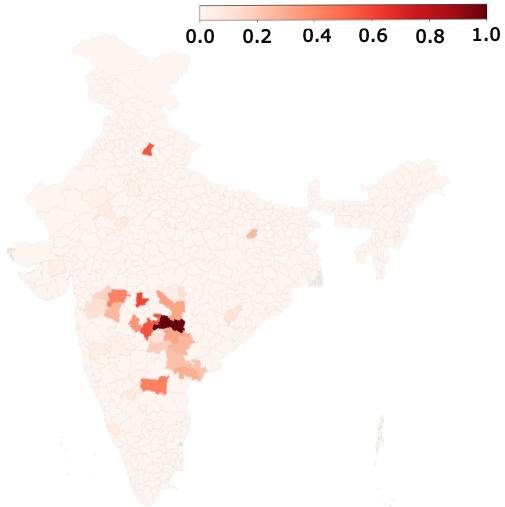}
    \caption{Bollworm on cotton queries during 2016.}
  \label{bollworm_news}
  \end{subfigure}
  \hfill
  \begin{subfigure}[b]{0.22\textwidth}
    \includegraphics[width=\textwidth]{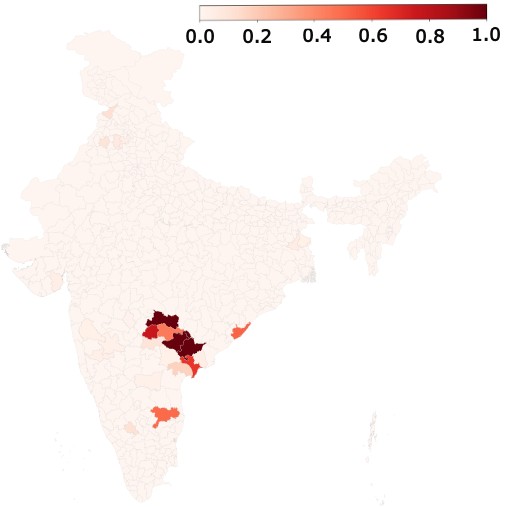}
    \caption{Armyworm on Maize during 2018.}
  \label{armyworm_news}
  \end{subfigure}
  \hfill
  \begin{subfigure}[b]{0.22\textwidth}
    \includegraphics[width=\textwidth]{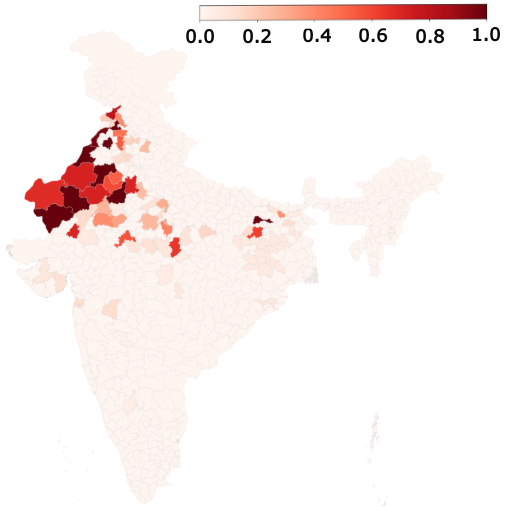}
    \caption{Locust infestation January 2020.}
  \label{locuts}
  \end{subfigure}
  \caption{Figure shows the normalized spatial distribution of the frequency of pest queries for whitefly, bollworm, armyworm, and locust.}
  \label{real_data}
\end{figure}
\vspace{-3mm}
\subsection{Similarity with real time events} 
In order to determine the veracity of our methodology for extracting useful information from queries, we compared the major pest attack events reported in popular news and reports to the spatio-temporal pest attack hot spots observed in our data. To this end, we scraped news articles from two of the most popular national newspapers, The Times of India (ToI) \cite{NewsLate1:online} and The Hindu \cite{TheHindu47:online} along with other sources such as DownToEarth. 

According to the report by DownToEarth \cite{Pestatta33:online}, whiteflies were responsible for immense crop loss during the year 2015-2016 in the states of Punjab and Haryana. We verified this report with our data by extracting whitefly queries for the year 2015-2016 as shown in Figure \ref{whitefly_news}. It shows that the states of Himachal Pradesh, Punjab, and Haryana, on average received the maximum number of queries per unit cultivated area. This information was also reported by The Hindu on August 25, 2015 \cite{Cottoncr16:online}, which was one month and 19 days later than the first query received in KCC. Similarly, our data display bollworm attack on Cotton (Kapas) during 2016 in the region of Andhra Pradesh and Maharashtra as shown in Figure \ref{bollworm_news} and the corresponding scenario was reported in \cite{bollwormnews}. In 2018, maize was reported to be affected by the armyworm in the districts of Andhra Pradesh \cite{FallArmy73:online}, and as expected, KCC also receives a large volume of maize related queries corresponding to armyworm from the regions of Andhra Pradesh shown in Figure \ref{armyworm_news}. It was first reported in the hindu \cite{Armyworm2:online} on October 08, 2018 which was 1 month 20 days later than the first query started appearing in the KCC. The much-publicized locust attack of 2020 also reported in  Wikipedia \cite{Locustsa24:online} in the northern part of India was observed in the queries related to locust attacks in regions of Rajasthan, Punjab, Himachal Pradesh as shown in Figure \ref{locuts}. It was also reported earlier in the KCC dataset. These results show crowdsourced data from KCC is helpful to analyze pest dynamics as it may be able to detect pest outbreaks earlier than reported in popular news papers.

\subsection{Pest Seasonality and Forecasting}
\subsubsection{Pest Seasonality}

\begin{figure}[!ht]
  \begin{subfigure}[b]{0.23\textwidth}
     \includegraphics[width=\linewidth]{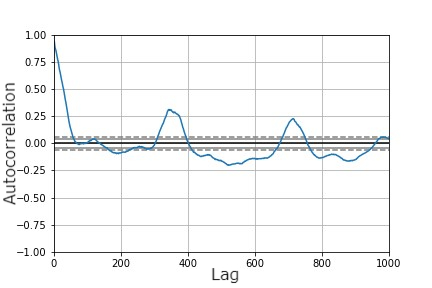}
  \caption{Whitefly}
  \label{at_whitefly}
  \end{subfigure}
  \hfill
  \begin{subfigure}[b]{0.23\textwidth}
    \includegraphics[width=\textwidth]{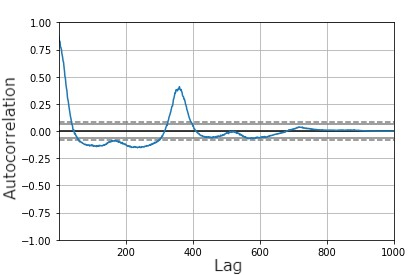}
    \caption{Bug}
  \label{at_bug}
  \end{subfigure}
  \hfill
  \begin{subfigure}[b]{0.23\textwidth}
     \includegraphics[width=\linewidth]{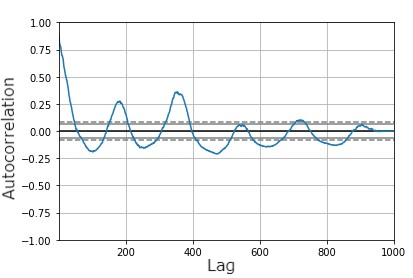}
  \caption{Aphid}
  \label{at_aphids}
  \end{subfigure}
  \hfill
  \begin{subfigure}[b]{0.23\textwidth}
    \includegraphics[width=\textwidth]{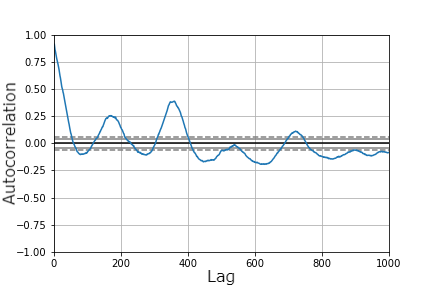}
    \caption{Stemborer}
  \label{at_stemborer}
  \end{subfigure}
  \caption{Figure shows auto-correlation of whitefly, bug, aphid and stemborer with minimum lag of 1 day. These plot indicate that pest show either yearly or half-year seasonality.}
  \label{auto_corr}
  \vspace{-3mm}
\end{figure}

We computed auto-correlation vs. lag plots of different pests with a minimum lag of a day. The auto-correlation plots of whitefly, bug, aphid and stemborer are shown in Figure \ref{auto_corr}.

\begin{figure}[!ht]
  \begin{subfigure}[b]{0.23\textwidth}
     \includegraphics[width=\linewidth]{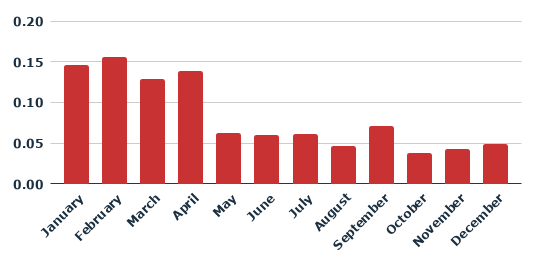}
  \caption{Whitefly}
  \label{whitefly_frequency}
  \end{subfigure}
  \hfill
  \begin{subfigure}[b]{0.23\textwidth}
    \includegraphics[width=\textwidth]{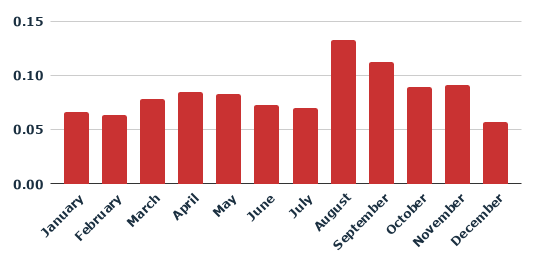}
    \caption{Bug}
  \label{bug_frequency}
  \end{subfigure}
  \hfill
  \begin{subfigure}[b]{0.23\textwidth}
     \includegraphics[width=\linewidth]{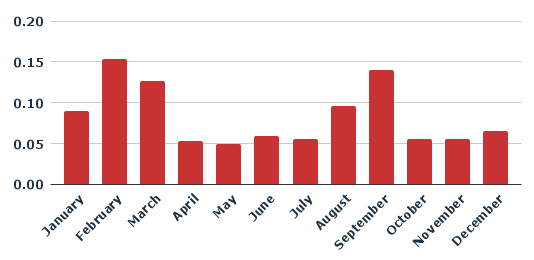}
  \caption{Aphid}
  \label{aphid_frequency}
  \end{subfigure}
  \hfill
  \begin{subfigure}[b]{0.23\textwidth}
    \includegraphics[width=\textwidth]{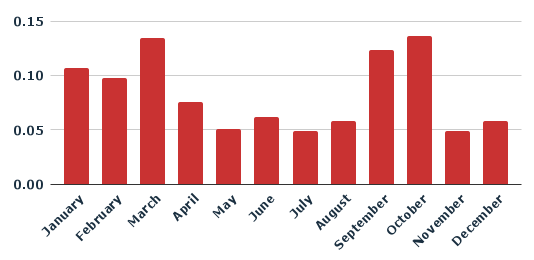}
    \caption{Stemborer}
  \label{stemborer_frequency}
  \end{subfigure}
  \caption{Figure shows frequency plot of whitefly, bug, aphid and stemborer over different months of year. These plots indicate pests are visible during specific times of the year.}
  \label{pest_fre}
  \vspace{-3mm}%
\end{figure}

 We observed peak in a time period of six month or an year. For whitefly, bug, aphid and stemborer over different months over five years is shown in Figure \ref{pest_fre}. Figure \ref{whitefly_frequency} and \ref{bug_frequency} shows white-fly and bug display yearly seasonality, where whitefly is observed during the months of January to April and bugs in August and September. Figure \ref{aphid_frequency} and \ref{stemborer_frequency} show aphid and stem borers reflect half-yearly seasonality and majorly found in the months of February, March and August, September.

\subsubsection{Forecasting}
We split our time series pest data into 70\% training and 30\% testing. We performed a DF test on the training data, and the results for aphid and termite are shown in Table \ref{table:dickey}. In the next step, we used grid search across the autoregressive (AR) and seasonally autoregressive (SAR) hyperparameters on training data to select best fitting parameters based on the lowest AIC value (as shown in Table \ref{table:model_parameter}). After setting the hyperparameters, a model is built by fitting the training data. Further, we used the trained model for prediction on the test data. Root mean square error, mean standard error and confidence interval corresponding to forecasting results of aphid, insect, bug, and stemborer are shown in Table \ref{tab:forecasting}

\begin{figure}[!ht]
  \begin{subfigure}[b]{0.23\textwidth}
  \centering
     \includegraphics[width=\linewidth]{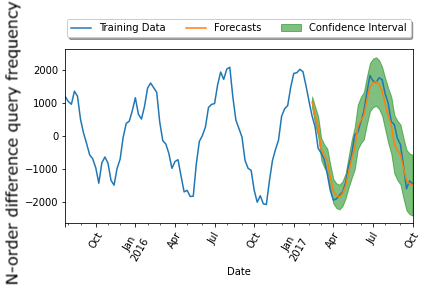}
  \caption{}
  \label{pre_aphid}
  \end{subfigure}
  \hfill
 \begin{subfigure}[b]{0.23\textwidth}
    \includegraphics[width=\textwidth]{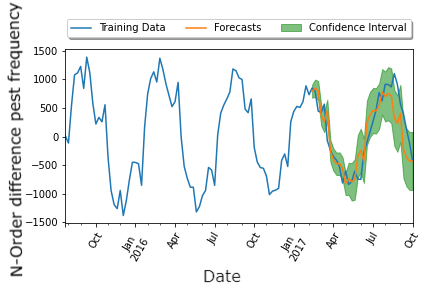}
    \caption{}
  \label{pre_termite}
  \end{subfigure}
  \begin{subfigure}[b]{0.23\textwidth}
  \centering
     \includegraphics[width=\linewidth]{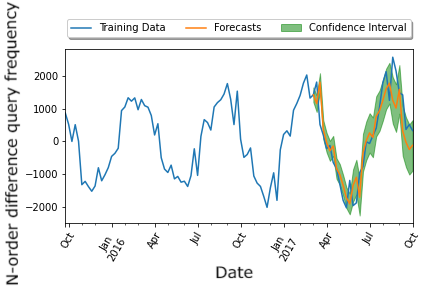}
  \caption{}
  \label{pre_aphid}
  \end{subfigure}
  \hfill
 \begin{subfigure}[b]{0.23\textwidth}
    \includegraphics[width=\textwidth]{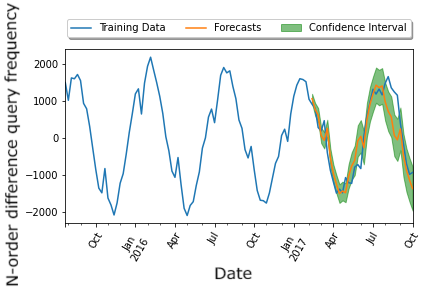}
    \caption{}
  \label{pre_termite}
  \end{subfigure}
  \caption{Figure shows the actual and forecast values by SARIMA model for aphid, insect, bug and stemborer using n-order frequency difference time series.}
  \label{forecasting}
  \vspace{5mm}
\end{figure}

\begin{table}[!ht]
\centering 
\begin{tabular}{|p{1.3cm}|p{1.5cm}|p{1.5cm}|p{2.8cm}|} 
\hline
\textbf{Pest} & \textbf{Mean Square Error} & \textbf{Mean Standard Error} & \textbf{Confidence Interval} \\ \hline
Aphid & 284.176 & 317.23 & [435.12, -808.415] \\ \hline
Insect & 313.369 & 172.860 & [390.48, -287.118] \\ \hline
Bug  & 313.56 & 172.35 & [302.56, -489.51] \\ \hline
Stemborer & 484.50 & 204.527 & [353.569, -448.166]  \\ \hline
\end{tabular}
\caption{Root mean square error, mean standard error and confidence interval for forecasting results of aphid, insect, bug, and stemborer} 
\label{tab:forecasting}
\vspace{-5mm}%
\end{table}


\section{Conclusion and Future work}

This study has drawn attention to the fact that crowdsourced data can be considered a way to study pest population dynamics as it mimics real-world events. Using this data, we investigated pest seasonality and found that their populations differ over time and place. We used time-series pest data to produce a reliable forecast, encouraging practical application. Our method can be used easily and provide an alternative approach to study pest attacks.In future, we intend to quantitatively validate our approach by possibly detecting all pest outbreak stories in the top news papers in a comprehensive manner. 

Early predictions of pest attacks can help us find effective ways to protect crops and achieve better harvests. This helps in knowing pest attack arrival time and identifies the period of occurrence. Thus forecasting the pest population will help farmers take appropriate action on a timely basis. It will also help farmers decide the amount of pesticides to be used, and it is desirable to plan appropriate control measures with maximum efficiency.


\bibliographystyle{ACM-Reference-Format}
\bibliography{pestbib}

\end{document}